\documentclass[letterpaper, 10 pt, conference]{ieeeconf}  

\IEEEoverridecommandlockouts                              
\usepackage[utf8]{inputenc}
\usepackage{multicol}
\usepackage[T1]{fontenc}
\usepackage[utf8]{inputenc}
\usepackage{authblk}
\usepackage{hyperref}
\usepackage{times}
\usepackage{latexsym}
\usepackage{biblatex}
\usepackage{url}
\usepackage{multicol}
\usepackage{booktabs}
\usepackage{graphicx}

\pdfoutput=1

\title{\LARGE \bf
ISS-MULT: Intelligent Sample Selection for\\Multi-Task Learning in Question Answering
}

\author{Ali Ahmadvand and Jinho D. Choi\\
  Computer Science,
  EMORY University,\\  
  Atlanta, GA 30322, USA\\
  {ali.ahmadvand@emory.edu} \\ 
  {jinho.choi@emory.edu} \\}

\begin{document}

\maketitle
\thispagestyle{empty}
\pagestyle{empty}

\begin{abstract}

Transferring knowledge from a source domain to another domain is useful, especially when gathering new data is very expensive and time-consuming. Deep networks have been well-studied for question answering tasks in recent years; however, no prominent research for transfer learning through deep neural networks exists in the question answering field. In this paper, two main methods (INIT and MULT) in this field are examined. Then, a new method named Intelligent sample selection (ISS-MULT) is proposed to improve the MULT method for question answering tasks. Different datasets, specificay SQuAD, SelQA, WikiQA, NewWikiQA and InforBoxQA, are used for evaluation. Moreover, two different tasks of question answering - answer selection and answer triggering - are evaluated to examine the effectiveness of transfer learning. The results show that using transfer learning generally improves the performance if the corpora are related and are based on the same policy. In addition, using ISS-MULT could finely improve the MULT method for question answering tasks, and these improvements prove more significant in the answer triggering task.

\end{abstract}

\section{INTRODUCTION}

Transferring knowledge between related domains could help to improve a learner on a special domain. Moreover, gathering data from different related resources proves very time-consuming and expensive. Therefore, this necessitates the development of machine learning methods to extract knowledge from these resources with different properties and characteristics. Transfer learning is a field in which machine learning methods encounter this challenge. 

Recently, deep neural networks outperform many machine learning methods in many real-world machine learning applications, especially NLP tasks. Consequently, deep learning methods have attracted much attention for question answering problems, and many methods based on deep learning algorithms have been proposed in recent years. Deep neural networks like other machine learning algorithms have some shortcomings. One of the main drawbacks of deep networks is their huge number of parameters which must be tuned during training. Consequently, deep neural networks require a huge amount of training samples to finely train and to prevent over-training. Also, using available and relevant data could improve their efficiency.
 
  In addition, although deep learning methods are well-studied for question answering tasks in recent years, there is no comprehensive research for transfer learning through deep neural networks in the question answering field. Deep networks are shallow for NLP applications compared to the computer vision tasks. Consequently, many ways to perform transfer learning in computer vision area are inapplicable to NLP tasks. 
 
 Mou at el. [6] conducted a comprehensive study on the transferability of the deep networks in sentence classification. They concluded that the best time to transfer the knowledge is when the tasks are semantically related because words are more abstract entities and convey more semantics than pixels in images. Moreover, fine-tuning the weights is more effective than freezing the weights during learning of the target dataset. Finally, MULT and INIT are generally comparable, and  Mou at el. did not observe any improvement by combining the two methods. They argue that the MULT method needs more in-depth
analysis in future work. Because of these challenges and the question answering gap not covered specifically, this paper performs transfer learning in question answering problems. Also, this paper follows Mou at el.'s recommendations and improves MULT using Intelligent Sample Selection (ISS) process. In the ISS, the most relevant samples to the source data is selected to transfer the knowledge.

\section{Related Works}

Transfer learning has a long history, and many researchers try to utilize the knowledge from relevant datasets to improve the performance on the target dataset [9, 7]. Moreover, transfer learning has been well-studied for deep learning methods in computer vision. Similarity between datasets is a key factor for performing transfer learning. Based on the similarity between two domains, different methods could be used. Bengio et. al [11] examine the ability of the neural networks in transferring knowledge from different domain. 

Transfer learning through deep networks is very restricted in NLP than Computer vision because the words naturally are high level entities, differing from the low level signals which create an image. However, there are some studies in the NLP realm which are restricted to sentence classification [6] and discourse relation classification [3]. Collobert and Weston [1] proposed a method in which they chose a random sample from both source and target domains with a $\lambda$ and 
$(1-\lambda)$ probability respectively, and subsequently the computed gradient is backpropagated through the network. Moreover, in Question answering, transfer learning through deep networks has not been well-studied because the networks are not deep like computer vision networks.

\subsection{Datasets}

Five different datasets (SQuAD, SelQA , WikiQA, WikiQA and InforbaxQA) are used for evaluation of the INIT, MULT and ISS-MULT methods. These datasets were proposed in recent years for the question answering problems. These datasets are produced differently. Therefore, they are may not be semanticaly related datasets, and this feature plays an important role in transfer learning in NLP tasks.

\textbf{SQuAD:} This dataset contains more than 107K crowdsourced questions on 536 Wikipedia article. SQuAD uses the entire candidate articles from Wikipedia to find the candidate answers. Also, the hit ratio for correct answers in SQuAD is about 35 percent [8].  

\textbf{SelQA:} This dataset is contains 8K+ questions in which more than half of the questions are paraphrased from the first half. In this dataset, like SQuAd and WikiQA, the entire articles from Wikipedia is searched to answer the questions [2].

\textbf{WikiQA:} This dataset includes questions selected from the Bing search queries. The abstract of articles in Wikipedia is used to find the answer candidates and the hit rate is about 36 percent. This dataset contains 20K+ question and answer. WikiQA differes from SQuAD and SelQA in its use of the abstract page of each article to find the candidate answers [10].

\textbf{WikiQA:} This dataset contains 9K+ question and answer pairs from original WikiQA. In WikiQA, unlike wikiQA, entire WikiPedia pages are search for candidate answers. Moreover, the hit rate (13\%) is much lower than the original WikiQA; therefore, this dataset is much more challenging than the WikiQA [13]. 

\textbf{InfoboxQA:} This dataset contains more than 15K questions extracted from 150 articles in Wikipedia. The answers lack complete sentence structure. Therefore, the nature of this dataset is differs from the other datasets [5].

\section{Methods}
sd
First, before mathematically discussing transfer learning, presenting a proper definition of this term is crucial. Generally, transfer learning is a process in which the knowledge from a special domain is transferred to a new domain with different properties of feature space and data distribution. This knowledge may be captured from learning a model which has already been trained on source data or through some mathematical mapping between two different domains. 

To mathematically define transfer learning we abide by the following notation [7]. Suppose each domain of \textit{D} consists of two main parts:
\begin{itemize}
\item The feature space $\chi$ with N-dimensions where $X = \{ x_1,...,x_n \} \in \chi$ 
\item The marginal distribution of $X, P(X)$.
\end{itemize}
For this domain, there exists a task named $T$ which also includes two main parts. The first part is the label space that refers to the label of the data $\gamma$, and the second part is the predictive function $f(X)$. Moreover, this function has been learned during learning process on the domain $D$. To summarize, for each transfer learning task, there is a domain $D = \{ \chi, P(X) \}$ and a task $T = \{ \gamma , f(X) \}$. Therefore, for transfer learning, it is a necessity to have at least two different domains and tasks. To this aim, consider $D_S$ as the source domain data where $D_S = \{ \chi_S,P(X_S) \}$. In the same way, $D_T$ is defined as the target domain data where $D_T = \{ \chi_T,P(X_T) \}$. Lastly, the source task is notated as $T_S = \{ \gamma_S,P(X_S) \}$, and the target task as $T_T = \{ \gamma_T,P(X_T) \}$. 

\subsection{INIT and Multi-Task Learning (MULT)}In the MULT method, two datasets are simultaneously trained, and the weights are tuned based on the inputs which come from both datasets. The hyper-parameter $\lambda \in (0,1)$ is calculated based on a brute-force search or using general global search. This hyper parameter is used to calculate the final cost function which is computed from the combination of the cost function of the source dataset and the target datasets. The final cost function is shown in Eq. 1.\\

$Cost = \lambda \times Cost(S) + (1-\lambda) \times Cos(T)$                (1)\\

where $S$ and $T$ are the source and target datasets respectively. One straightforward way to compute the cost function is randomly select the samples from both datasets with the probability of $\lambda$ and then compute the cost function for each sample. This parameter is used to randomly select a sample from a special dataset during the training. 

The way that INIT and MULT method transfer knowledge varies drastically. In INIT, a initial point for optimization process is estimated instead of a random point selection. This initialization is very tricky for gradient based methods; however, sometimes this does not work in complex feature spaces of the problem. On the other hand, in MULT the samples from two dataset simultaneously affect the optimization process in which the source dataset behaves like a regularizer and potentially prevents the network from overfitting on the data.

\begin{table*}[h]
\centering
\begin{tabular}{l||lll||lll||lll}
\multicolumn{10}{c}{\bf{Evaluation On}} \\

\multicolumn{1}{c||}{\textbf{Trained On}}&
\multicolumn{3}{c}{WikiQA} & 
\multicolumn{3}{c}{SelQA} &
\multicolumn{3}{c}{SQuAD} \\

\cline{2-10}
\multicolumn{1}{c||}{} &\multicolumn{1}{c}{MAP} & \multicolumn{1}{c}{MRR} & \multicolumn{1}{c||}{F1} & \multicolumn{1}{c}{MAP} & \multicolumn{1}{c}{MRR} & \multicolumn{1}{c||}{F1} & \multicolumn{1}{c}{MAP} & \multicolumn{1}{c}{MRR} & \multicolumn{1}{c}{F1} \\
\hline\hline
WikiQA & 65.54& 67.41& 13.33  & 82.99&84.04&45.63 & 89.71&88.87&-  \\
SelQA & \bf{66.57} & \bf{68.23} & \bf{16.67}  &  82.72&83.70 & 46.49 & \bf{89.80} & \bf{88.97} & - \\
SQuAD &   65.83 & 67.50 & 15.09 & \bf{83.41} & \bf{84.04}& \bf{47.63} & 89.51&88.64& -  \\
InfoBox & - & -& - & 82.27 & 83.31 & 40.77 & - & - & -  \\ 
\end{tabular}
\caption{Results for MULT.}
\label{my-label}
\end{table*}

\subsection{ISS-MULT}
The MULT method needs more research in NLP [6]. One possible improvement of this method is to automatically tune the hyperparameter of lambda during training in which a unique and proper lambda has been calculated for a special dataset. Although it seems that the estimation of a proper lambda is not a trivial work, there are some restrictions which help users to choose a proper lambda. First of all, the range of the lambda is predetermined and is between 0.85 and 1 because the network needs to see more data from the target dataset in each epoch. In addition, the lambda's behavior on mean average precision (MAP), mean reciprocal rank (MRR) and F1-score is very complex. In other words, there are many local optima in this multi-objective optimization problem. Moreover, there is not much difference between global optimum and the other local optima in this range. 

Another way to improve this method could be to select the samples which are more relevant to the target dataset. Based on the importance of the similarity between the datasets for transfer learning in the NLP tasks, this paper proposes to use the most relevant samples from the source dataset to train on the target dataset. One way to find the most similar samples is finding the pair-wise distance between all samples of the development set of the target dataset and source dataset.

This idea encounters two main problems. First, in our experiments the source dataset is a huge dataset like SQuAD with more than 107K samples. Second, comparing two questiona and answer pairs using the cosine similarity will not be a fast task, especially when each word is represented in a vector of length 300.

To solve this problem, we propose using a clustering algorithm on the development set. The clustering algorithm used ihere is a hierarchical clustering algorithm. The cosine similarity is used as a criteria to cluster each question and answer. Therefore, these clusters are representative of the development set of the target dataset and the corresponding center for each cluster is representative of all the samples on that cluster. In the next step, the distance of each center is used to calculate the cosine similarity. Finally, the samples in the source dataset which are far from these centers are ignored. In other words, the outliers do not take part in transfer learning.

\section{Experiments}

For evaluation of INIT, MULT and ISS-MULT method, the bigram CNN introduced by Yu et al. [12] is used. This model consists of two 2D-convolution layers and each convolution layer is followed by a pooling layer. GoogleNews-word2vec is used as a pre-trained embedding layer to represent each word. Forty different filters of size $(2 \times embedding size (300))$ are used to create the feature spaces for each layer. Then, a logistic regression is used to predict the final result. 

In this paper, two main question answering tasks such as answer selection and answer triggering have been examined. In the answer triggering task, there is not a guarantee to have the correct answer among the list of answers. However, in answer selection, there is at least one correct answer among the candidates. As a result, answer triggering is a more challenging task. To report the result for answer selection, MAP and MRR are used; however, the answer triggering task is evaluated by F1-score. The result for MULT Method is reported in Table. 1.

The results have shown that the MULT method could properly improve all metrics for both answer selection and answer triggering task. Moreover, this improvement is more significant in answer triggering. The result for InforboxQA dataset shows that this dataset deteriorate all metrics because the nature of this dataset is different from the other datasets. Moreover, the improvement on SQuAD is not significant because this dataset is a big dataset and already contains enough data to fine-tune the parameters of the deep network. 

 In other experiment, the INIT method is implemented. In the INIT method, the weights for the best results on the development set of the source dataset is used for initialization of the deep network. The results for the INIT method is listed in Table 2. Moreover, The results for ISS-MULT method is shown in Table 3.

\begin{table*}[h]
\centering
\begin{tabular}{@{}l||lll||lll@{}}
\multicolumn{7}{c}{\bf{Evaluation On}} \\ 

\multicolumn{1}{c||}{\textbf{Trained On}}&
\multicolumn{3}{c}{WikiQA} & 
\multicolumn{3}{c}{SelQA} \\

\cline{2-7}

\multicolumn{1}{c||}{} & \multicolumn{1}{c}{MAP} & \multicolumn{1}{c}{MRR} & \multicolumn{1}{c||}{F1} & \multicolumn{1}{c}{MAP} & \multicolumn{1}{c}{MRR} & \multicolumn{1}{c}{F1} \\
\hline\hline
WikiQA & 65.54& 67.41& 13.33  & 83.44& 84.45& \bf{46.59}  \\
SelQA &   66.48& 68.48& \bf{17.14}  &  82.72&83.70& 46.49 \\
SQuAD &  \bf{ 66.89} & \bf{69.03} & 16.33 & \bf{83.77} & \bf{84.82} & 46.44  \\
\end{tabular}
\caption{Results for INIT}
\label{my-label}
\end{table*}

\begin{table*}
\centering
\begin{tabular}{@{}l||lll||lll@{}}
\multicolumn{7}{c}{\bf{Evaluation On}} \\ 
\multicolumn{1}{c||}{\textbf{Trained On}}&
\multicolumn{3}{c}{WikiQA} & 
\multicolumn{3}{c}{SelQA} \\

\cline{2-7}
\multicolumn{1}{c||}{} & \multicolumn{1}{c}{MAP} & \multicolumn{1}{c}{MRR} & \multicolumn{1}{c||}{F1} & \multicolumn{1}{c}{MAP} & \multicolumn{1}{c}{MRR} & \multicolumn{1}{c}{F1} \\
\hline\hline
WikiQA & 65.54 & 67.41 & 13.33  & 83.25 & 84.27 & 47.00  \\
SelQA  &  67.08 & 69.23 & 20.83  &  82.72 & 83.70 & 46.49 \\
SQuAD &   \bf{67.12} & \bf{68.77} & \bf{23.73} & \bf{83.40} & \bf{84.37} & \bf{47.73} \\
\end{tabular}
\caption{Results for ISS-MULT}
\label{my-label}
\end{table*}

 The results indicate that the ISS-MULT method could improve all metrics on both the WikiQA and SelQA datasets. This improvement is more obvious in answer triggering task. 
 
 We also performe another experiment to examine INIT and MULT method for original WikiQA. The F1-score for this dataset is equal to 33.73; however, the average INIT result for both SQuAD and SelQA as initializers is 30.50. In addition, the average results for MULT and ISS-MULT  are 31.76 and 32.65, respectively.
 The result on original WikiQA indicates that all three transfer learning methods not only do not improve the results but also hurt the F1-score. Therefore, SelQA and SQuAD could not estimate a proper initial point for gradient based optimization method. Moreover, these corpora could not refine the error surface of the original WikiQA dataset during optimization for MULT and ISS-MULT method. 
 
 These are because other datasets could not add new information to the original dataset or they apparently add some redundant information which are dissimilar to the target dataset. Although ISS-MULT tries to remove this effect and consequently the result is improved, this method is on top of MULT method, and the result is significantly based on the effectiveness of this method.
 
 According to Section 3, SQuAD, SelQA and WikiQA are created based on the same policy from entire Wikipedia pages; however, the original WikiQA is just created based on the abstract of each page. Therefore, the original WikiQA dataset is different from WikiQA, SelQA and SQuAD most likely because INIT and MULT method do not work for original WikiQA based on available corpora. 
 
 All in all, indicated Table. 1, 2 and 3, the INIT method slightly generates better results for the answer selection task compared to MULT and ISS-MULT . Moreover, all of the three methods improve the MAP and MRR metrics compared to the base method. On the other hand, MULT and ISS-MULT produce much better results for answer triggering task than the INIT method. In this task, all three methods outperform the base method. Moreover, according to our experiments, using different policies to generate different datasets could intensely affect transfer learning for the answer triggering task.

\section{Conclusions}

In this paper, we presented a comprehensive experiment for two main methods in deep learning on five recent corpora for question answering and triggering tasks. A new method on the top of MULT named ISS-MULT is presented. The results show that transfer learning could generally improve the results and this improvement is larger in the answer triggering task. According to the results, we reach the conclusion that transfer learning works best on semantically related corpora but also works well on datasets created similarly.

\section{Acknowledgement}
The authors would like to thank Derek Onken, Massimiliano Lupo Pasini and Tomasz Jurczyk for their help in providing datasets and their guidance in this paper.

\end{document}